\documentclass[
]{ceurart}

\usepackage{listings}
\usepackage[most]{tcolorbox}
\begin{document}

\copyrightyear{2022}
\copyrightclause{Copyright for this paper by its authors.
  Use permitted under Creative Commons License Attribution 4.0
  International (CC BY 4.0).}

\conference{GeoExT Workshop, ECIR 2025}

\title{Generating Synthetic Oracle Datasets to Analyze Noise Impact: A Study on Building Function Classification Using Tweets}





\author[1,3]{Shanshan Bai}[
email=shanshan.bai@tum.de,
orcid=0009-0000-4786-8809
]
\author[2]{Anna Kruspe}[
orcid=0000-0002-2041-9453
]
\author[1,3]{Xiaoxiang Zhu}[
orcid=0000-0001-5530-3613
]

\address[1]{Technical University of Munich}
\address[2]{Munich University of Applied Sciences}
\address[3]{Munich Center for Machine Learning}


\begin{abstract}
Tweets provides valuable semantic context for earth observation tasks and serves as a complementary modality to remote sensing imagery. In building function classification (BFC), tweets are often collected using geographic heuristics and labeled via external databases—an inherently weakly supervised process that introduces both label noise and sentence-level feature noise (e.g., irrelevant or uninformative tweets). While label noise has been widely studied, the impact of sentence-level feature noise remains underexplored, largely due to the lack of clean benchmark datasets for controlled analysis. In this work, we propose a method for generating a synthetic oracle dataset using LLM, designed to contain only tweets that are both correctly labeled and semantically relevant to their associated buildings. This oracle dataset enables systematic investigation of noise impacts that are otherwise difficult to isolate in real-world data. To assess its utility, we compare model performance using Naïve Bayes and mBERT classifiers under three configurations: real vs. synthetic training data, and cross-domain generalization. Results show that noise in real tweets significantly degrades the contextual learning capacity of mBERT, reducing its performance to that of a simple keyword-based model. In contrast, the clean synthetic dataset allows mBERT to learn effectively, outperforming Naïve Bayes by a large margin. These findings highlight that addressing feature noise is more critical than model complexity in this task. Our synthetic dataset offers a novel experimental environment for future noise injection studies and is publicly available on GitHub\footnote{\url{https://github.com/zhu-xlab/synthetic_tweets}}.
\end{abstract}


\maketitle

\section{Introduction}\label{sec:Introduction}
Building Function Classification (BFC) aims to determine the functional purpose of buildings, such as commercial or residential use, using various data sources. Traditionally, remote sensing imagery has been the dominant modality for this task. However, such imagery often lacks the semantic granularity needed to distinguish nuanced urban functions at the building level, for example, differentiating between dormitories and hotels.

Textual data from social media, such as geo-tagged tweets, offers a complementary perspective. Tweets sent from the same location as a building may reveal human activities and behaviors that provide contextual clues about building use. For instance, a tweet like "\textit{Great coffee at this café!}" implies a commercial function, while "\textit{Silent night at the dorm.}" suggests residential use.

Despite this promise, social media datasets collected for BFC often rely on weak supervision—tweets are heuristically matched to buildings (e.g. tweets send within 50 meters radius of a building are assigned to this building) and labeled using voluntarily-provided building tags on the external databases such as OpenStreetMap (OSM). This process introduces multiple types of noise: (1) label noise, from incorrect or outdated building tags; and (2) sentence-level feature noise, in the form of irrelevant, ambiguous, or uninformative tweets. While label noise has been extensively studied—often through controlled label-flipping experiments—sentence-level feature noise remains harder to investigate. This is because it requires access to a dataset that is known to contain only relevant and correctly labeled examples, something that human annotators often cannot guarantee due to the subjective and implicit nature of semantic relevance.

To address this, we explore the feasibility of using LLM to generate a synthetic oracle dataset for BFC: a noise-free benchmark in which all tweets are both correctly labeled and semantically relevant to their associated buildings. We define an oracle dataset as an idealized collection designed not for deployment, but to serve as a clean experimental environment for systematically analyzing the effects of noise. 

It is also important to acknowledge that geo-tagged tweets are no longer widely accessible due to platform changes (e.g., Twitter removing precise geolocation tagging \cite{kruspe2021changes}). However, our focus is not on Twitter per se, but on the broader challenge of noise in weakly supervised user-generated datasets—an issue that persists across many social media applications.

Our key contributions are as follows:
\begin{description}
\item [A Synthetic Oracle Tweets Dataset for Building Function Classification:] We construct a clean, LLM-generated dataset that enables controlled experiments on feature noise, facilitating analysis that is otherwise infeasible using real-world data.
\item [A Data Generation Pipeline:] We introduce a reproducible method for generating synthetic datasets, guided by real-world building and tweet distributions to ensure statistical realism.
\item [Evaluation of Data Quality:] We compare the correctness and diversity of the synthetic data against real-world datasets using both classifier performance and linguistic metrics (Self-BLEU, perplexity).
\item [Insights into Feature Noise Impact:] Our results show that handling feature noise is more critical than increasing model complexity for BFC tasks.
\end{description}

The remainder of this paper is structured as follows: Section \ref{sec:Related Work} reviews related work on building function classification and synthetic data generation. Section \ref{sec:Methodologies} describes our data generation pipeline. Section \ref{sec:Evaluation} presents the evaluation methodology and results. Section \ref{sec:Discussions} analyzes the quality and characteristics of the dataset. Section \ref{sec:Conclusion} summarizes our findings, and Section \ref{sec:Future Work} outlines directions for future research.

\section{Related Work}\label{sec:Related Work}

\subsection*{Building Function Classification with Tweets}

Recent studies have investigated the feasibility of using geo-tagged tweets for Building Function Classification (BFC), confirming their potential while emphasizing key challenges related to noise and data quality. \citet{34} were among the first to treat BFC as a text classification task, assigning tweets located near buildings as inputs. However, their use of FastText limited robustness in multilingual urban settings. In contrast, \citet{35} shifted focus from sentence-level classification to feature engineering, using tweet embeddings to extract function-indicative features. While some keywords were found to correlate strongly with building types, overlapping lexical features introduced noise, reducing overall reliability.

To enhance classification accuracy, \citet{36} proposed a decision-level fusion strategy, integrating textual features from tweets with remote sensing imagery. Their results showed that while tweets offered useful contextual cues, the discriminative power largely came from imagery. More recently, \citet{dissertation} scaled BFC globally, reframing it as a multilingual classification task using mBERT. Despite achieving moderate success (55\% accuracy across three functional categories), noise in the tweet data—both in labeling and content—remained a significant bottleneck.

A consistent finding across these studies is that tweets, while informative, are inherently noisy. Label noise often originates from incorrect or incomplete tags in OpenStreetMap (OSM) \cite{labelnoise,15}, and geo-tagging mismatch occurs when tweets are attributed to the wrong building due to spatial heuristics \cite{24}. Additionally, sentence-level feature noise—such as irrelevant weets—remains a major challenge \cite{22}. While label noise has been the subject of many experimental studies, there has been limited systematic analysis of sentence-level feature noise and its effects on model performance.

\subsection*{Synthetic Data Generation with LLMs}

Large Language Models (LLMs) have shown impressive ability to generate text that captures patterns and structures from real-world corpora \cite{3}. Recent work has explored using LLMs to create synthetic datasets via instruction-tuned prompting conditioned on class labels \cite{4,5}, showing that LLMs can serve as scalable alternatives to human annotators. However, category-conditioned generation poses a critical challenge: the synthetic text distribution often diverges from real data in linguistic diversity, topical focus, or context relevance \cite{30,31}.

To address these issues, researchers have proposed human-in-the-loop refinement strategies \cite{7} and model-guided feedback loops \cite{29}, which adjust prompts based on classifier performance. Others have explored persona-based synthesis \cite{8,9,10,11} to introduce style and perspective variation. While these approaches enhance fluency and diversity, they are not explicitly designed for tasks requiring spatial and contextual grounding—such as aligning tweets with building-specific metadata in urban environments.

Despite these advances, there remains a gap in applying LLM-based data generation to fine-grained, geospatially anchored classification tasks like BFC. Most prior work assumes structured label spaces and ignores the semantic constraints imposed by geographic or functional context. Our work addresses this gap by conditioning LLM prompts on real-world building attributes (e.g., function, name, location) and matching tweet language distributions. This enables the creation of a high-quality oracle dataset that supports controlled analysis of noise in text-based BFC.

\section{Synthetic Tweets Generation Pipeline} \label{sec:Methodologies}

This section presents our three-step pipeline for generating synthetic multilingual tweets using a large language model (LLM), conditioned on building-level metadata. The pipeline includes: (1) retrieving metadata, (2) cleaning it, and (3) prompting the LLM to generate contextually grounded tweets.

\subsection{Step 1: Retrieving Metadata}

We define metadata as structured descriptive information associated with each building—such as its function, location, and the intended number and language of tweets to be generated. This metadata is either inherited from a prior labeled dataset \cite{dissertation} or retrieved from OSM, ensuring that the resulting synthetic dataset reflects the statistical characteristics of real-world data.
\begin{itemize}
\item \textbf{Building Attributes:}
\begin{itemize}
   \item \texttt{Building\_tag}: A fine-grained functional label from OSM (e.g., \texttt{"restaurant"}, \texttt{"apartment"}), distinct from the binary ground-truth classification ("commercial" or "residential"). 
   \item \texttt{Building\_name}: A descriptive identifier used in tweet content (e.g., "Merlex Auto Group"). 
   \item \texttt{Building\_city}: The city where the building is located. 
\end{itemize}
\item \textbf{Tweets Language Distribution:}
A list specifying the language of each synthetic tweet to be generated. For instance, \texttt{["English", "English"]} indicates two English tweets should be generated for that building. These values are inherited from the dataset in \cite{dissertation}, ensuring alignment with real-world tweet frequency and language usage. 
\end{itemize}

\subsection{Step 2: Preprocessing Metadata}

Since building attributes derived from OSM can be noisy, and we also want to control the maximum number of tweets each building, we apply a cleaning preprocessing pipeline to previous collected metadata. This step ensures LLM generating synthetic tweets without introducing unwanted noise and also a balanced dataset.

\begin{itemize} 

\item Removing formatting artifacts: We sanitize building names and tags by stripping special characters (e.g., underscores \_, slashes \/) to prevent malformed prompts.

\item Filtering generic or erroneous tags: Entries with non-informative tags such as `"yes"` or `"roof"` are excluded, as they do not convey meaningful function.

\item Ensuring unique tag: Many buildings are associated with multiple use-specific tags that suggest different functional categories (e.g., both `"church"` and `"restaurant"`). 

\item Ensuring label-tag consistency: For buildings pre-labeled as `"commercial"` or `"residential"` (from prior datasets), we remove those whose OSM tags clearly contradict the assigned label. For example, if a building labeled `"commercial"` has the tag `"mosque"`, it is removed from the dataset. 

\item Restricting to at most five tweets per building: This step applies to the number of synthetic tweets to be generated and also to real-world tweets. We cap the number at five to balance the dataset across buildings and to control for prompt length and generation consistency. This helps avoid overrepresentation of a few buildings during training or evaluation phases.

\end{itemize}

The preprocessed metadata is stored in structured JSON format. An example is shown below:

\begin{lstlisting}[language=Python]
{"Building city": "WashingtonDC",
 "Building tag": "Retail",
 "Building name": "Merlex Auto Group",
 "Tweets distribution": ["English", "English"]}
\end{lstlisting}

While these preprocessing steps helps reduce ambiguity and maintain consistency, we acknowledge that it may exclude buildings with complex, multi-functional roles (e.g., shopping malls with food courts, theaters, and offices). Although this limits the coverage of our analysis, our focus in this study is to establish a controlled, noise-free experimental setting—not to comprehensively model all real-world function types.

\subsection{Step 3: Generating Tweets using an LLM}
We use the \texttt{Llama-3.3-70B-Instruct-bnb-4bit} model from Hugging Face\footnote{\url{https://huggingface.co/unsloth/Llama-3.3-70B-Instruct-bnb-4bit}} to generate multilingual tweets. Each prompt consists of two key components:
\begin{itemize}
    \item System Prompt: Defines the overall task, outlining style and formatting guidelines for tweet generation. The system prompt also includes a one-shot example demonstrating the desired tweet format. This remains constant for all buildings.
    \item User Prompt: Contains building-specific metadata, to ensure diverse and contextually relevant outputs.
\end{itemize}
Here is the system prompt:
\begin{tcolorbox}[colback=white!95!gray, colframe=black, boxrule=0.5mm, arc=2mm, width=\textwidth]
\textit{Generate tweets as if they were posted by real Twitter users in a specific building. Tweets should be sent from the type of building described in the 'building tag'. Ensure that each tweet reflects a unique perspective or experience. Imagine switching personas for each tweet, simulating thoughts from different types of users, such as tourists, professionals, or families. Consider varying the tone (e.g., humorous, cynical, formal, casual), the length (short and concise, or longer and detailed), and the use of mentions or hashtags. Highlight varied aspects of the building, such as its architecture, services, location, history, or events. You must generate only one tweet in each language specified under 'tweet language distribution,' written directly in that language.}\\\\
\textit{Example:}

\textit{\{"Building\_city": "WashingtonDC",\\
"Building\_tag": "Retail",\\
"Building\_name": "Merlex Auto Group",\\
"Tweets\_language\_distribution": ["English", "English"]\}\\}

\textit{["Bought new rims here at Merlex Auto yesterday, totally transformed my ride! @merlexautogroup \#AutoCare", "Merlex Auto Group really knows how to treat car lovers right. The staff? Super knowledgeable. The selection? If you’re in DC and thinking about upgrading your ride, this is the place! \#CarShopping \#DCLuxury"]}
\end{tcolorbox}

Using this three-steps pipeline, we generate a synthetic oracle dataset covering 6,000 real-world buildings across 41 global cities. The final dataset includes 15,222 tweets in 45 languages. We ensure that the distribution of building types, tweet language frequencies closely mirror those of the real-world dataset used in \cite{dissertation}. A quantitative comparison between real and synthetic tweets is provided in Table \ref{tab:real_synthetic}.

\begin{table}[h!]
    \centering
    \renewcommand{\arraystretch}{1.6} 
    \setlength{\tabcolsep}{8pt}       
    \small 
    \resizebox{\linewidth}{!}{ 
        \begin{tabular}{|p{1.8cm}|p{7cm}|p{7.2cm}|} 
            \hline
            \rowcolor[gray]{0.9} \textbf{Category} & \textbf{Commercial} & \textbf{Residential} \\
            \hline
            \textbf{Tag} & Retail & Dormitory \\ 
            \hline
            \textbf{Name} & Superfresh & Baxter Hall \\
            \hline
            \textbf{City} & New York & Cape Town \\
            \hline
            \textbf{Tweet Distribution} & 
            \texttt{[English, English, English]} & 
            \texttt{[English, English, English, English, English]} \\
            \hline
            \textbf{Real Tweets} &  
            \begin{minipage}[t]{7cm}
                \raggedright
                "Just posted a photo @ SuperFresh Supermarket of Bay Ridge." \\ \vspace{5pt} 
                "Turkey Dinner with all the Fixin's, that your kids will actually gobble up?! Yep! Cake \#nailedit Spotted at superfresh\_bayridge\_dyker (where they do NOT make vegan pumpkin pie btw) \@ SuperFresh Bay Ridge." \\ \vspace{5pt} 
                "New York - The food in the deli, had to go to the hospital where they pumped my stomach out and determined the deli salads poisoned me. I was so... Food Poisoning."
            \end{minipage} &  
            \begin{minipage}[t]{7.2cm}
                \raggedright
                "Just posted a photo @ Baxter Hall." \\ \vspace{5pt} 
                "Happiness is a decision @ Baxter Hall." \\ \vspace{5pt} 
                "Just posted a video @ Baxter Hall." \\ \vspace{5pt} 
                "Shooting visserlieb clarinet improv \#artistic \#research \#clarinet \#performancebasedresearch \#phdlife \#phdstudent \#archive \#freeimprovisation \#kirbycollection \#music…." \\ \vspace{5pt} 
                "Shooting visserlieb clarinet improv. \#artistic \#research \#clarinet \#artisticresearch \#performancebasedresearch \#phdlife \#phdstudent \#archive \#freeimprovisation \#kirbycollection \#music \#experimentation \#sound…."
            \end{minipage} \\
            \hline
            \textbf{Synthetic Tweets} &  
            \begin{minipage}[t]{7cm}
                \raggedright
                "Fresh produce at unbeatable prices, Superfresh has become my go-to grocery spot in New York!" \\ \vspace{5pt} 
                "Why did I wait so long to discover the best deals on electronics are actually inside this store?" \\ \vspace{5pt} 
                "Does anyone else notice how vibrant the flower arrangements are near the entrance every morning? It sets the mood right for shopping."
            \end{minipage} &  
            \begin{minipage}[t]{7.2cm}
                \raggedright
                "Cramming for exams is way more bearable when you have an amazing view like ours at Baxter Hall!" \\ \vspace{5pt} 
                "Dorm life can be crazy but our community here makes all the difference." \\ \vspace{5pt} 
                "Battling to get work done before deadline hits - anyone else feeling the pressure? Hope caffeine kicks in soon." \\ \vspace{5pt} 
                "Sharing laughs over breakfast on the rooftop has become daily highlight always something new." \\ \vspace{5pt} 
                "Wonderful place full great people making every day count."
            \end{minipage} \\
            \hline
        \end{tabular}
    }
    \caption{Comparison of Real-World and Synthetic Tweets of Two Buildings}
    \label{tab:real_synthetic}
\end{table}

\section{Dataset Evaluations and Results} \label{sec:Evaluation}

In this section, we evaluate the quality of the generated synthetic dataset along two key dimensions: diversity and correctness. Diversity ensures that the dataset captures a broad range of sentence structures and vocabulary variations, rather than overly repetitive content that could oversimplify the classification task. Correctness assesses whether the synthetic dataset fulfills its intended purpose as an oracle dataset, containing only perfectly aligned tweets that semantically correspond to their respective target buildings.

Since each building in the synthetic dataset mirrors a real-world building exactly, and the tweet distributions in terms of volume and language match their real-world counterparts, we evaluate correctness and diversity by comparing the synthetic dataset against the real-world dataset.

\subsection{Diversity Evaluation}

To assess diversity, we use 4-gram Self-BLEU \cite{selfBLEU} as the primary metric, following \cite{32}. Self-BLEU measures how similar each sentence is to the rest of the dataset by calculating BLEU \cite{selfBLEU} scores for every sentence against all others. A lower Self-BLEU score indicates higher diversity, suggesting that the dataset contains a richer variety of expressions. The results are reported in Table \ref{tab:diversity}.

As a secondary measure, we compute perplexity \cite{17}. While traditionally used to assess how well a language model predicts text, perplexity can also serve as an indirect proxy for vocabulary alignment between the synthetic and real-world datasets. Specifically, we compute unigram perplexity using a unigram language model pre-trained on a held-out set of 100,000 real-world tweets from buildings not included in the experimental dataset. By evaluating perplexity for both datasets, a similar score between synthetic and real-world data suggests that the synthetic dataset shares comparable linguistic distribution and vocabulary complexity. Conversely, a significant deviation would indicate substantial differences in word usage. The perplexity results are also included in Table \ref{tab:diversity}.

\begin{table*}[htbp]
\centering
\begin{tabular}{lcc}
\toprule
\textbf{Dataset} & \textbf{Self-BLEU}$\downarrow$ & \textbf{Perplexity(\( \log_{10} \))}$\downarrow$ \\
\midrule
Synthetic    & 48.37\%   & 4.49 \\
Real-world    & 40.78\%   & 4.36   \\
\bottomrule
\end{tabular}
\caption{Diversity Comparison Results by Self-BLEU and Perplexity on Synthetic and Real-world Datasets}
\label{tab:diversity}
\end{table*}

\subsection{Correctness Evaluation}

\begin{table*}[htbp]
\centering
\setlength{\tabcolsep}{8pt} 
\renewcommand{\arraystretch}{1.2} 
\small 
\resizebox{\textwidth}{!}{%
\begin{tabular}{@{}p{3cm}lcccccc@{}c@{}}
\toprule 
\textbf{Configurations} & \textbf{Model} & \multicolumn{3}{c}{\textbf{Commercial}} & \multicolumn{3}{c}{\textbf{Residential}} & \textbf{Overall Accuracy} \\
\cmidrule(lr){3-5} \cmidrule(lr){6-8} 
 & & Precision & Recall & F1 & Precision & Recall & F1 & \\
\midrule 
\multirow{2}{3cm}{Real-world} 
& NB      & 0.67 $\pm$ 0.03 & 0.57 $\pm$ 0.02 & 0.61 $\pm$ 0.02 & 0.61 $\pm$ 0.02 & 0.71 $\pm$ 0.02 & 0.66 $\pm$ 0.02 & 0.64 $\pm$ 0.02 \\
& mBERT   & 0.66 $\pm$ 0.01 & 0.68 $\pm$ 0.03 & 0.67 $\pm$ 0.02 & 0.66 $\pm$ 0.02 & 0.62 $\pm$ 0.01 & 0.64 $\pm$ 0.01 & 0.65 $\pm$ 0.01 \\
\midrule
\multirow{2}{3cm}{Synthetic} 
& NB       & 0.86 $\pm$ 0.02 & 0.82 $\pm$ 0.03 & 0.84 $\pm$ 0.02 & 0.81 $\pm$ 0.03 & 0.85 $\pm$ 0.01 & 0.83 $\pm$ 0.02 & 0.84 $\pm$ 0.02 \\
& mBERT   & 0.92 $\pm$ 0.03 & 0.90 $\pm$ 0.01 & 0.91 $\pm$ 0.01 & 0.90 $\pm$ 0.01 & 0.92 $\pm$ 0.02 & 0.91 $\pm$ 0.01 & 0.91 $\pm$ 0.01 \\
\midrule
\multirow{2}{3cm}{Cross-Domain} 
& NB       & 0.60 $\pm$ 0.03 & 0.67 $\pm$ 0.01 & 0.63 $\pm$ 0.02 & 0.60 $\pm$ 0.02 & 0.52 $\pm$ 0.02 & 0.56 $\pm$ 0.02 & 0.60 $\pm$ 0.01 \\
& mBERT   & 0.59 $\pm$ 0.06 & 0.77 $\pm$ 0.06 & 0.66 $\pm$ 0.01 & 0.66 $\pm$ 0.02 & 0.44 $\pm$ 0.15 & 0.52 $\pm$ 0.10 & 0.61 $\pm$ 0.05 \\
\bottomrule 
\end{tabular}%
}
\caption{Performance comparison of Naïve Bayes (NB) and mBERT models across three configurations: Real-world (train and test on real-world data), Synthetic (traid and test on synthetic data), and Cross-Domain (train on synthetic data and test on real-world data).}
\label{tab:accuracy}
\end{table*}

A direct way to evaluate correctness is by testing the dataset’s utility in a downstream classification task. We compare two distinct classifiers: Naïve Bayes (NB) and multilingual BERT (mBERT). These models were chosen to contrast a traditional statistical classifier (NB), which relies on word co-occurrence patterns, with a transformer-based language model (mBERT), which leverages contextual embeddings for nuanced language understanding. We conduct evaluations under three configurations:

\begin{itemize}
    \item \textbf{Real-world:} Train and test models using real-world data to establish a baseline.
    \item \textbf{Synthetic:} Train and test models using synthetic data to determine if the dataset achieves correctness.
    \item \textbf{Cross-domain:} Train models on synthetic data and test on real-world data to evaluate generalization performance.
\end{itemize}

First two configurations allow us to assess whether the synthetic dataset can function as a valid oracle dataset. The cross-domain evaluation is particularly insightful in determining how well models trained on noise-free synthetic data adapt to noisy real-world conditions.

The real-world dataset used in our experiments was derived from \cite{dissertation}, filtered to include the same 6,000 buildings as the synthetic set. For each building, we retain the same number of tweets in the same languages, resulting in a dataset of 15,222 real tweets. We split both datasets at the building level to prevent data leakage, allocating 10,729 tweets from 80\% of buildings for training and validation, and 2279 tweets of the rest 20\% buildings for testing. We fine-tune \texttt{bert-base-multilingual-cased} for five epochs using the \texttt{Adam optimizer} with a learning rate of 5e-6, a batch size of 16, a warm-up ratio of 0.01, and a weight decay of 0.01 to prevent overfitting. All training hyperparameters remain consistent across experiments. The results are presented in Table \ref{tab:accuracy}.

\section{Discussions} \label{sec:Discussions}

Our diversity evaluation reveals a trade-off between structural variation and lexical realism in the synthetic dataset. The higher self-BLEU score of the synthetic tweets (48.37\%) compared to the real-world tweets (40.78\%) suggests that the synthetic content is more repetitive at the sentence structure level. However, the perplexity scores—4.49 for synthetic vs. 4.36 for real—indicate that the vocabulary distribution is comparable across both datasets. This suggests that while the LLM may favor certain patterns or templates in sentence generation, it still captures a realistic range of word usage and topic-relevant terms.

In the real-world configuration, mBERT (65\%) performs only marginally better than Naïve Bayes (64\%). This finding suggests that the contextual capabilities of transformer models are suppressed by the high level of noise in real tweets. Naïve Bayes, which relies on surface-level word co-occurrences, performs nearly as well—indicating that complex models may be overkill in noisy, weakly supervised settings where semantic signals are diluted.

By contrast, in the synthetic configuration, mBERT (91\%) significantly outperforms Naïve Bayes (84\%). This performance gap highlights how the noise-free, semantically aligned tweets in the oracle dataset allow transformer model to leverage its full potential. The performance gain achieved by mBERT also reflects the presence of fine-grained semantic cues that word-frequency based models cannot exploit. This validates the intended function of our synthetic dataset as an oracle environment—models capable of contextual understanding should perform well when noise is removed. 

These findings also show a key insight: noise handling is more critical than model complexity for improving classification accuracy in weakly supervised text settings. Even the best models fail to learn effectively when irrelevant or ambiguous inputs dominate the data.

We also noted the cross-domain evaluation shows that models trained on the synthetic oracle dataset generalize poorly to noisy real-world tweets. This result underscores the significant impact of domain shift introduced by the removal of noise by LLM-generated dataset. It is important to recognize that our synthetic dataset is not designed to replace real-world data. Rather, it serves a distinct purpose: to create a controlled, noise-free environment for systematic experimentation—particularly for studying the isolated effects of label or feature noise via noise injection strategies. We also acknowledge the potential risk of semantic drift in LLM-generated data—that is, the possibility that synthetic tweets may reflect biases, stereotypes, or linguistic abstractions learned by the model rather than replicating authentic user behavior. However, our goal is not to reproduce social media behavior with full fidelity. Instead, we intentionally trade off some realism for semantic clarity and control, enabling cleaner experimental analysis.

\section{Conclusion} \label{sec:Conclusion} 

This study investigates the feasibility of using LLM to generate a synthetic oracle dataset for text classification tasks. We focus on the domain of building function classification (BFC) using geo-tagged tweets, a modality that offers semantic richness but suffers from label noise and, more critically, sentence-level feature noise—irrelevant, ambiguous, or uninformative tweets that obscure learning signals.

While label noise has been extensively studied through noise injection experiments, sentence-level feature noise remains underexplored due to the lack of a truly clean dataset. Human annotators often disagree on what constitutes relevance in social media text, making it difficult to filter out such noise consistently. To address this, we construct an oracle dataset using an LLM, where all tweets are guaranteed to be correctly labeled and semantically aligned with their associated building types. This provides a controlled, noise-free environment suitable for systematic experimentation.

Our evaluations show that:
\begin{itemize}
    \item The synthetic dataset, while slightly less diverse in sentence structure (higher Self-BLEU), exhibits comparable lexical richness (similar perplexity) to real-world tweets.
    \item On real-world data, sophisticated models like mBERT perform only marginally better than simple classifiers like Naïve Bayes, due to noise degrading contextual learning.
    \item On synthetic data, mBERT significantly outperforms Naïve Bayes, validating that the synthetic dataset retains meaningful semantic distinctions and fulfills its oracle role.
\end{itemize}

These findings suggest that addressing feature noise may be more critical than increasing model complexity for tasks involving weakly supervised text. While our synthetic dataset is not intended to replicate real user behavior, it offers a valuable testbed for controlled experiments that would be infeasible with noisy real-world data.

\section{Future Work}\label{sec:Future Work}

One key challenge in synthetic data generation is balancing diversity and semantic fidelity. Our findings show that while our dataset maintains realistic vocabulary usage, its sentence structure is less varied than real-world tweets. Future work could explore techniques such as controlled paraphrasing, persona variation, or prompt augmentation to increase structural diversity without compromising label alignment. Reducing this domain gap would make synthetic datasets more useful not only for experimentation but also for training and transfer learning.

Our synthetic oracle dataset provides a clean, well-defined starting point for studying the impact of noise in text classification. Future research can build on this by systematically injecting different types and levels of noise—such as label flipping, irrelevant content, or off-topic language—to isolate their individual and combined effects on model performance. 

Given that our synthetic tweets are grounded in real cities and building names, future work could use this dataset to examine how geospatial context influences text classification. For example, studies could compare performance with and without location-specific references, or investigate how city-level variation affects model generalization. This has implications not only for BFC, but also for broader geo-aware NLP tasks such as place-based sentiment analysis or urban event detection.


\section*{References}
\begin{thebibliography}{23}
\expandafter\ifx\csname natexlab\endcsname\relax\def\natexlab#1{#1}\fi
\providecommand{\url}[1]{\texttt{#1}}
\providecommand{\href}[2]{#2}
\providecommand{\path}[1]{#1}
\providecommand{\DOIprefix}{doi:}
\providecommand{\ArXivprefix}{arXiv:}
\providecommand{\URLprefix}{URL: }
\providecommand{\Pubmedprefix}{pmid:}
\providecommand{\doi}[1]{\href{http://dx.doi.org/#1}{\path{#1}}}
\providecommand{\Pubmed}[1]{\href{pmid:#1}{\path{#1}}}
\providecommand{\bibinfo}[2]{#2}
\ifx\xfnm\relax \def\xfnm[#1]{\unskip,\space#1}\fi
\bibitem[{Kruspe et~al.(2021)Kruspe, H{\"a}berle, Hoffmann, Rode-Hasinger,
  Abdulahhad, and Zhu}]{kruspe2021changes}
\bibinfo{author}{A.~Kruspe}, \bibinfo{author}{M.~H{\"a}berle},
  \bibinfo{author}{E.~J. Hoffmann}, \bibinfo{author}{S.~Rode-Hasinger},
  \bibinfo{author}{K.~Abdulahhad}, \bibinfo{author}{X.~X. Zhu},
\newblock \bibinfo{title}{Changes in {T}witter geolocations: Insights and
  suggestions for future usage},
\newblock in: \bibinfo{editor}{W.~Xu}, \bibinfo{editor}{A.~Ritter},
  \bibinfo{editor}{T.~Baldwin}, \bibinfo{editor}{A.~Rahimi} (Eds.),
  \bibinfo{booktitle}{Proceedings of the Seventh Workshop on Noisy
  User-generated Text (W-NUT 2021)}, \bibinfo{publisher}{Association for
  Computational Linguistics}, \bibinfo{address}{Online}, \bibinfo{year}{2021},
  pp. \bibinfo{pages}{212--221}. \URLprefix
  \url{https://aclanthology.org/2021.wnut-1.24/}.
  \DOIprefix\doi{10.18653/v1/2021.wnut-1.24}.
\bibitem[{Häberle et~al.(2019{\natexlab{a}})Häberle, Werner, and Zhu}]{34}
\bibinfo{author}{M.~Häberle}, \bibinfo{author}{M.~Werner},
  \bibinfo{author}{X.~X. Zhu},
\newblock \bibinfo{title}{Building type classification from social media texts
  via geo-spatial textmining}  (\bibinfo{year}{2019}{\natexlab{a}})
  \bibinfo{pages}{10047--10050}. \URLprefix
  \url{https://ieeexplore.ieee.org/document/8898836}.
\bibitem[{Häberle et~al.(2019{\natexlab{b}})Häberle, Werner, and Zhu}]{35}
\bibinfo{author}{M.~Häberle}, \bibinfo{author}{M.~Werner},
  \bibinfo{author}{X.~X. Zhu},
\newblock \bibinfo{title}{Geo-spatial text-mining from twitter – a feature
  space analysis with a view toward building classification in urban regions},
\newblock \bibinfo{journal}{European Journal of Remote Sensing}
  \bibinfo{volume}{52} (\bibinfo{year}{2019}{\natexlab{b}})
  \bibinfo{pages}{2–11}. \URLprefix
  \url{https://doi.org/10.1080/22797254.2019.1586451}.
  \DOIprefix\doi{10.1080/22797254.2019.1586451}.
\bibitem[{Häberle et~al.(2022)Häberle, Hoffmann, and Zhu}]{36}
\bibinfo{author}{M.~Häberle}, \bibinfo{author}{E.~J. Hoffmann},
  \bibinfo{author}{X.~X. Zhu},
\newblock \bibinfo{title}{Can linguistic features extracted from geo-referenced
  tweets help building function classification in remote sensing?},
\newblock \bibinfo{journal}{ISPRS Journal of Photogrammetry and Remote Sensing}
  \bibinfo{volume}{188} (\bibinfo{year}{2022}) \bibinfo{pages}{255--268}.
  \URLprefix
  \url{https://www.sciencedirect.com/science/article/pii/S0924271622001058}.
  \DOIprefix\doi{https://doi.org/10.1016/j.isprsjprs.2022.04.006}.
\bibitem[{Häberle(2022)}]{dissertation}
\bibinfo{author}{M.~Häberle},
\newblock \bibinfo{title}{Fusion of remote sensing images and social media text
  messages for building function classification}  (\bibinfo{year}{2022})
  \bibinfo{pages}{134}. \URLprefix \url{https://mediatum.ub.tum.de/1637448}.
\bibitem[{Frenay and Verleysen(2014)}]{labelnoise}
\bibinfo{author}{B.~Frenay}, \bibinfo{author}{M.~Verleysen},
\newblock \bibinfo{title}{Classification in the presence of label noise: A
  survey},
\newblock \bibinfo{journal}{IEEE Transactions on Neural Networks and Learning
  Systems} \bibinfo{volume}{25} (\bibinfo{year}{2014})
  \bibinfo{pages}{845--869}. \DOIprefix\doi{10.1109/TNNLS.2013.2292894}.
\bibitem[{Gu et~al.(2024)Gu, Chen, Shi, and Zhu}]{15}
\bibinfo{author}{Z.~Gu}, \bibinfo{author}{Z.~Chen}, \bibinfo{author}{Y.~Shi},
  \bibinfo{author}{X.~Zhu},
\newblock \bibinfo{title}{Building attributes recognition with noisy and
  incomplete labels},
\newblock \bibinfo{year}{2024}, pp. \bibinfo{pages}{230--233}.
  \DOIprefix\doi{10.1109/IGARSS53475.2024.10641987}, \bibinfo{note}{publisher
  Copyright: {\textcopyright} 2024 IEEE.; 2024 IEEE International Geoscience
  and Remote Sensing Symposium, IGARSS 2024 ; Conference date: 07-07-2024
  Through 12-07-2024}.
\bibitem[{Lamsal et~al.(2022)Lamsal, Harwood, and Read}]{24}
\bibinfo{author}{R.~Lamsal}, \bibinfo{author}{A.~Harwood},
  \bibinfo{author}{M.~R. Read},
\newblock \bibinfo{title}{Where did you tweet from? inferring the origin
  locations of tweets based on contextual information},
\newblock in: \bibinfo{booktitle}{2022 IEEE International Conference on Big
  Data (Big Data)}, \bibinfo{publisher}{IEEE}, \bibinfo{year}{2022}, p.
  \bibinfo{pages}{3935–3944}. \URLprefix
  \url{http://dx.doi.org/10.1109/BigData55660.2022.10020460}.
  \DOIprefix\doi{10.1109/bigdata55660.2022.10020460}.
\bibitem[{Kruspe et~al.(2021)Kruspe, Kersten, and Klan}]{22}
\bibinfo{author}{A.~Kruspe}, \bibinfo{author}{J.~Kersten},
  \bibinfo{author}{F.~Klan},
\newblock \bibinfo{title}{Review article: Detection of actionable tweets in
  crisis events},
\newblock \bibinfo{journal}{Natural Hazards and Earth System Sciences}
  \bibinfo{volume}{21} (\bibinfo{year}{2021}) \bibinfo{pages}{1825--1845}.
  \URLprefix \url{https://nhess.copernicus.org/articles/21/1825/2021/}.
  \DOIprefix\doi{10.5194/nhess-21-1825-2021}.
\bibitem[{Delétang et~al.(2024)Delétang, Ruoss, Duquenne, Catt, Genewein,
  Mattern, Grau-Moya, Wenliang, Aitchison, Orseau, Hutter, and Veness}]{3}
\bibinfo{author}{G.~Delétang}, \bibinfo{author}{A.~Ruoss},
  \bibinfo{author}{P.-A. Duquenne}, \bibinfo{author}{E.~Catt},
  \bibinfo{author}{T.~Genewein}, \bibinfo{author}{C.~Mattern},
  \bibinfo{author}{J.~Grau-Moya}, \bibinfo{author}{L.~K. Wenliang},
  \bibinfo{author}{M.~Aitchison}, \bibinfo{author}{L.~Orseau},
  \bibinfo{author}{M.~Hutter}, \bibinfo{author}{J.~Veness},
  \bibinfo{title}{Language modeling is compression}, \bibinfo{year}{2024}.
  \URLprefix \url{https://arxiv.org/abs/2309.10668}.
  \href{http://arxiv.org/abs/2309.10668}{{\tt arXiv:2309.10668}}.
\bibitem[{Ge et~al.(2024)Ge, Hu, Wang, Wang, Chen, and Wei}]{4}
\bibinfo{author}{T.~Ge}, \bibinfo{author}{J.~Hu}, \bibinfo{author}{L.~Wang},
  \bibinfo{author}{X.~Wang}, \bibinfo{author}{S.-Q. Chen},
  \bibinfo{author}{F.~Wei}, \bibinfo{title}{In-context autoencoder for context
  compression in a large language model}, \bibinfo{year}{2024}. \URLprefix
  \url{https://arxiv.org/abs/2307.06945}.
  \href{http://arxiv.org/abs/2307.06945}{{\tt arXiv:2307.06945}}.
\bibitem[{Schick and Sch{\"u}tze(2021)}]{5}
\bibinfo{author}{T.~Schick}, \bibinfo{author}{H.~Sch{\"u}tze},
\newblock \bibinfo{title}{Generating datasets with pretrained language models},
\newblock in: \bibinfo{editor}{M.-F. Moens}, \bibinfo{editor}{X.~Huang},
  \bibinfo{editor}{L.~Specia}, \bibinfo{editor}{S.~W.-t. Yih} (Eds.),
  \bibinfo{booktitle}{Proceedings of the 2021 Conference on Empirical Methods
  in Natural Language Processing}, \bibinfo{publisher}{Association for
  Computational Linguistics}, \bibinfo{address}{Online and Punta Cana,
  Dominican Republic}, \bibinfo{year}{2021}, pp. \bibinfo{pages}{6943--6951}.
  \URLprefix \url{https://aclanthology.org/2021.emnlp-main.555/}.
  \DOIprefix\doi{10.18653/v1/2021.emnlp-main.555}.
\bibitem[{Brown et~al.(2020)Brown, Mann, Ryder, Subbiah, Kaplan, Dhariwal,
  Neelakantan, Shyam, Sastry, Askell, Agarwal, Herbert-Voss, Krueger, Henighan,
  Child, Ramesh, Ziegler, Wu, Winter, Hesse, Chen, Sigler, Litwin, Gray, Chess,
  Clark, Berner, McCandlish, Radford, Sutskever, and Amodei}]{30}
\bibinfo{author}{T.~B. Brown}, \bibinfo{author}{B.~Mann},
  \bibinfo{author}{N.~Ryder}, \bibinfo{author}{M.~Subbiah},
  \bibinfo{author}{J.~Kaplan}, \bibinfo{author}{P.~Dhariwal},
  \bibinfo{author}{A.~Neelakantan}, \bibinfo{author}{P.~Shyam},
  \bibinfo{author}{G.~Sastry}, \bibinfo{author}{A.~Askell},
  \bibinfo{author}{S.~Agarwal}, \bibinfo{author}{A.~Herbert-Voss},
  \bibinfo{author}{G.~Krueger}, \bibinfo{author}{T.~Henighan},
  \bibinfo{author}{R.~Child}, \bibinfo{author}{A.~Ramesh},
  \bibinfo{author}{D.~M. Ziegler}, \bibinfo{author}{J.~Wu},
  \bibinfo{author}{C.~Winter}, \bibinfo{author}{C.~Hesse},
  \bibinfo{author}{M.~Chen}, \bibinfo{author}{E.~Sigler},
  \bibinfo{author}{M.~Litwin}, \bibinfo{author}{S.~Gray},
  \bibinfo{author}{B.~Chess}, \bibinfo{author}{J.~Clark},
  \bibinfo{author}{C.~Berner}, \bibinfo{author}{S.~McCandlish},
  \bibinfo{author}{A.~Radford}, \bibinfo{author}{I.~Sutskever},
  \bibinfo{author}{D.~Amodei}, \bibinfo{title}{Language models are few-shot
  learners}, \bibinfo{year}{2020}. \URLprefix
  \url{https://arxiv.org/abs/2005.14165}.
  \href{http://arxiv.org/abs/2005.14165}{{\tt arXiv:2005.14165}}.
\bibitem[{Bommasani et~al.(2022)Bommasani, Hudson, Adeli, Altman, Arora, von
  Arx, Bernstein, Bohg, Bosselut, Brunskill, Brynjolfsson, Buch, Card,
  Castellon, Chatterji, Chen, Creel, Davis, Demszky, Donahue, Doumbouya,
  Durmus, Ermon, Etchemendy, Ethayarajh, Fei-Fei, Finn, Gale, Gillespie, Goel,
  Goodman, Grossman, Guha, Hashimoto, Henderson, Hewitt, Ho, Hong, Hsu, Huang,
  Icard, Jain, Jurafsky, Kalluri, Karamcheti, Keeling, Khani, Khattab, Koh,
  Krass, Krishna, Kuditipudi, Kumar, Ladhak, Lee, Lee, Leskovec, Levent, Li,
  Li, Ma, Malik, Manning, Mirchandani, Mitchell, Munyikwa, Nair, Narayan,
  Narayanan, Newman, Nie, Niebles, Nilforoshan, Nyarko, Ogut, Orr,
  Papadimitriou, Park, Piech, Portelance, Potts, Raghunathan, Reich, Ren, Rong,
  Roohani, Ruiz, Ryan, Ré, Sadigh, Sagawa, Santhanam, Shih, Srinivasan,
  Tamkin, Taori, Thomas, Tramèr, Wang, Wang, Wu, Wu, Wu, Xie, Yasunaga, You,
  Zaharia, Zhang, Zhang, Zhang, Zhang, Zheng, Zhou, and Liang}]{31}
\bibinfo{author}{R.~Bommasani}, \bibinfo{author}{D.~A. Hudson},
  \bibinfo{author}{E.~Adeli}, \bibinfo{author}{R.~Altman},
  \bibinfo{author}{S.~Arora}, \bibinfo{author}{S.~von Arx},
  \bibinfo{author}{M.~S. Bernstein}, \bibinfo{author}{J.~Bohg},
  \bibinfo{author}{A.~Bosselut}, \bibinfo{author}{E.~Brunskill},
  \bibinfo{author}{E.~Brynjolfsson}, \bibinfo{author}{S.~Buch},
  \bibinfo{author}{D.~Card}, \bibinfo{author}{R.~Castellon},
  \bibinfo{author}{N.~Chatterji}, \bibinfo{author}{A.~Chen},
  \bibinfo{author}{K.~Creel}, \bibinfo{author}{J.~Q. Davis},
  \bibinfo{author}{D.~Demszky}, \bibinfo{author}{C.~Donahue},
  \bibinfo{author}{M.~Doumbouya}, \bibinfo{author}{E.~Durmus},
  \bibinfo{author}{S.~Ermon}, \bibinfo{author}{J.~Etchemendy},
  \bibinfo{author}{K.~Ethayarajh}, \bibinfo{author}{L.~Fei-Fei},
  \bibinfo{author}{C.~Finn}, \bibinfo{author}{T.~Gale},
  \bibinfo{author}{L.~Gillespie}, \bibinfo{author}{K.~Goel},
  \bibinfo{author}{N.~Goodman}, \bibinfo{author}{S.~Grossman},
  \bibinfo{author}{N.~Guha}, \bibinfo{author}{T.~Hashimoto},
  \bibinfo{author}{P.~Henderson}, \bibinfo{author}{J.~Hewitt},
  \bibinfo{author}{D.~E. Ho}, \bibinfo{author}{J.~Hong},
  \bibinfo{author}{K.~Hsu}, \bibinfo{author}{J.~Huang},
  \bibinfo{author}{T.~Icard}, \bibinfo{author}{S.~Jain},
  \bibinfo{author}{D.~Jurafsky}, \bibinfo{author}{P.~Kalluri},
  \bibinfo{author}{S.~Karamcheti}, \bibinfo{author}{G.~Keeling},
  \bibinfo{author}{F.~Khani}, \bibinfo{author}{O.~Khattab},
  \bibinfo{author}{P.~W. Koh}, \bibinfo{author}{M.~Krass},
  \bibinfo{author}{R.~Krishna}, \bibinfo{author}{R.~Kuditipudi},
  \bibinfo{author}{A.~Kumar}, \bibinfo{author}{F.~Ladhak},
  \bibinfo{author}{M.~Lee}, \bibinfo{author}{T.~Lee},
  \bibinfo{author}{J.~Leskovec}, \bibinfo{author}{I.~Levent},
  \bibinfo{author}{X.~L. Li}, \bibinfo{author}{X.~Li}, \bibinfo{author}{T.~Ma},
  \bibinfo{author}{A.~Malik}, \bibinfo{author}{C.~D. Manning},
  \bibinfo{author}{S.~Mirchandani}, \bibinfo{author}{E.~Mitchell},
  \bibinfo{author}{Z.~Munyikwa}, \bibinfo{author}{S.~Nair},
  \bibinfo{author}{A.~Narayan}, \bibinfo{author}{D.~Narayanan},
  \bibinfo{author}{B.~Newman}, \bibinfo{author}{A.~Nie}, \bibinfo{author}{J.~C.
  Niebles}, \bibinfo{author}{H.~Nilforoshan}, \bibinfo{author}{J.~Nyarko},
  \bibinfo{author}{G.~Ogut}, \bibinfo{author}{L.~Orr},
  \bibinfo{author}{I.~Papadimitriou}, \bibinfo{author}{J.~S. Park},
  \bibinfo{author}{C.~Piech}, \bibinfo{author}{E.~Portelance},
  \bibinfo{author}{C.~Potts}, \bibinfo{author}{A.~Raghunathan},
  \bibinfo{author}{R.~Reich}, \bibinfo{author}{H.~Ren},
  \bibinfo{author}{F.~Rong}, \bibinfo{author}{Y.~Roohani},
  \bibinfo{author}{C.~Ruiz}, \bibinfo{author}{J.~Ryan},
  \bibinfo{author}{C.~Ré}, \bibinfo{author}{D.~Sadigh},
  \bibinfo{author}{S.~Sagawa}, \bibinfo{author}{K.~Santhanam},
  \bibinfo{author}{A.~Shih}, \bibinfo{author}{K.~Srinivasan},
  \bibinfo{author}{A.~Tamkin}, \bibinfo{author}{R.~Taori},
  \bibinfo{author}{A.~W. Thomas}, \bibinfo{author}{F.~Tramèr},
  \bibinfo{author}{R.~E. Wang}, \bibinfo{author}{W.~Wang},
  \bibinfo{author}{B.~Wu}, \bibinfo{author}{J.~Wu}, \bibinfo{author}{Y.~Wu},
  \bibinfo{author}{S.~M. Xie}, \bibinfo{author}{M.~Yasunaga},
  \bibinfo{author}{J.~You}, \bibinfo{author}{M.~Zaharia},
  \bibinfo{author}{M.~Zhang}, \bibinfo{author}{T.~Zhang},
  \bibinfo{author}{X.~Zhang}, \bibinfo{author}{Y.~Zhang},
  \bibinfo{author}{L.~Zheng}, \bibinfo{author}{K.~Zhou},
  \bibinfo{author}{P.~Liang}, \bibinfo{title}{On the opportunities and risks of
  foundation models}, \bibinfo{year}{2022}. \URLprefix
  \url{https://arxiv.org/abs/2108.07258}.
  \href{http://arxiv.org/abs/2108.07258}{{\tt arXiv:2108.07258}}.
\bibitem[{Schick and Schütze(2021)}]{7}
\bibinfo{author}{T.~Schick}, \bibinfo{author}{H.~Schütze},
  \bibinfo{title}{Generating datasets with pretrained language models},
  \bibinfo{year}{2021}. \URLprefix \url{https://arxiv.org/abs/2104.07540}.
  \href{http://arxiv.org/abs/2104.07540}{{\tt arXiv:2104.07540}}.
\bibitem[{Ouyang et~al.(2022)Ouyang, Wu, Jiang, Almeida, Wainwright, Mishkin,
  Zhang, Agarwal, Slama, Ray, Schulman, Hilton, Kelton, Miller, Simens, Askell,
  Welinder, Christiano, Leike, and Lowe}]{29}
\bibinfo{author}{L.~Ouyang}, \bibinfo{author}{J.~Wu},
  \bibinfo{author}{X.~Jiang}, \bibinfo{author}{D.~Almeida},
  \bibinfo{author}{C.~L. Wainwright}, \bibinfo{author}{P.~Mishkin},
  \bibinfo{author}{C.~Zhang}, \bibinfo{author}{S.~Agarwal},
  \bibinfo{author}{K.~Slama}, \bibinfo{author}{A.~Ray},
  \bibinfo{author}{J.~Schulman}, \bibinfo{author}{J.~Hilton},
  \bibinfo{author}{F.~Kelton}, \bibinfo{author}{L.~Miller},
  \bibinfo{author}{M.~Simens}, \bibinfo{author}{A.~Askell},
  \bibinfo{author}{P.~Welinder}, \bibinfo{author}{P.~Christiano},
  \bibinfo{author}{J.~Leike}, \bibinfo{author}{R.~Lowe},
  \bibinfo{title}{Training language models to follow instructions with human
  feedback}, \bibinfo{year}{2022}. \URLprefix
  \url{https://arxiv.org/abs/2203.02155}.
  \href{http://arxiv.org/abs/2203.02155}{{\tt arXiv:2203.02155}}.
\bibitem[{Shanahan et~al.(2023)Shanahan, McDonell, and Reynolds}]{8}
\bibinfo{author}{M.~Shanahan}, \bibinfo{author}{K.~McDonell},
  \bibinfo{author}{L.~Reynolds}, \bibinfo{title}{Role-play with large language
  models}, \bibinfo{year}{2023}. \URLprefix
  \url{https://arxiv.org/abs/2305.16367}.
  \href{http://arxiv.org/abs/2305.16367}{{\tt arXiv:2305.16367}}.
\bibitem[{Li et~al.(2024)Li, Mehrabi, Peris, Goyal, Chang, Galstyan, Zemel, and
  Gupta}]{9}
\bibinfo{author}{J.~Li}, \bibinfo{author}{N.~Mehrabi},
  \bibinfo{author}{C.~Peris}, \bibinfo{author}{P.~Goyal},
  \bibinfo{author}{K.-W. Chang}, \bibinfo{author}{A.~Galstyan},
  \bibinfo{author}{R.~Zemel}, \bibinfo{author}{R.~Gupta}, \bibinfo{title}{On
  the steerability of large language models toward data-driven personas},
  \bibinfo{year}{2024}. \URLprefix \url{https://arxiv.org/abs/2311.04978}.
  \href{http://arxiv.org/abs/2311.04978}{{\tt arXiv:2311.04978}}.
\bibitem[{Ge et~al.(2024)Ge, Chan, Wang, Yu, Mi, and Yu}]{10}
\bibinfo{author}{T.~Ge}, \bibinfo{author}{X.~Chan}, \bibinfo{author}{X.~Wang},
  \bibinfo{author}{D.~Yu}, \bibinfo{author}{H.~Mi}, \bibinfo{author}{D.~Yu},
  \bibinfo{title}{Scaling synthetic data creation with 1,000,000,000 personas},
  \bibinfo{year}{2024}. \URLprefix \url{https://arxiv.org/abs/2406.20094}.
  \href{http://arxiv.org/abs/2406.20094}{{\tt arXiv:2406.20094}}.
\bibitem[{Choi and Li(2024)}]{11}
\bibinfo{author}{H.~K. Choi}, \bibinfo{author}{Y.~Li}, \bibinfo{title}{Picle:
  Eliciting diverse behaviors from large language models with persona
  in-context learning}, \bibinfo{year}{2024}. \URLprefix
  \url{https://arxiv.org/abs/2405.02501}.
  \href{http://arxiv.org/abs/2405.02501}{{\tt arXiv:2405.02501}}.
\bibitem[{Zhu et~al.(2018)Zhu, Lu, Zheng, Guo, Zhang, Wang, and Yu}]{selfBLEU}
\bibinfo{author}{Y.~Zhu}, \bibinfo{author}{S.~Lu}, \bibinfo{author}{L.~Zheng},
  \bibinfo{author}{J.~Guo}, \bibinfo{author}{W.~Zhang},
  \bibinfo{author}{J.~Wang}, \bibinfo{author}{Y.~Yu}, \bibinfo{title}{Texygen:
  A benchmarking platform for text generation models}, \bibinfo{year}{2018}.
  \URLprefix \url{https://arxiv.org/abs/1802.01886}.
  \href{http://arxiv.org/abs/1802.01886}{{\tt arXiv:1802.01886}}.
\bibitem[{Ye et~al.(2022)Ye, Gao, Li, Xu, Feng, Wu, Yu, and Kong}]{32}
\bibinfo{author}{J.~Ye}, \bibinfo{author}{J.~Gao}, \bibinfo{author}{Q.~Li},
  \bibinfo{author}{H.~Xu}, \bibinfo{author}{J.~Feng}, \bibinfo{author}{Z.~Wu},
  \bibinfo{author}{T.~Yu}, \bibinfo{author}{L.~Kong}, \bibinfo{title}{Zerogen:
  Efficient zero-shot learning via dataset generation}, \bibinfo{year}{2022}.
  \URLprefix \url{https://arxiv.org/abs/2202.07922}.
  \href{http://arxiv.org/abs/2202.07922}{{\tt arXiv:2202.07922}}.
\bibitem[{Jurafsky(2025)}]{17}
\bibinfo{author}{D.~Jurafsky}, \bibinfo{title}{Speech and Language Processing},
  \bibinfo{publisher}{Stanford University}, \bibinfo{year}{2025}.

\end{thebibliography}

\end{document}